\documentclass[conference,letterpaper]{IEEEtran}
\IEEEoverridecommandlockouts
\usepackage{cite}
\usepackage{amsmath,amssymb,amsfonts}
\allowdisplaybreaks[4]
\usepackage[ruled]{algorithm2e}
\usepackage{multirow}
\usepackage{booktabs}
\usepackage{textcomp}
\usepackage{xcolor}
\usepackage{subcaption}
\usepackage{graphicx}
\usepackage{makecell}
\usepackage{verbatim} 
\usepackage{url}
\usepackage{threeparttable}
\usepackage{bbding}
\usepackage{hyperref}

\def\BibTeX{{\rm B\kern-.05em{\sc i\kern-.025em b}\kern-.08em
    T\kern-.1667em\lower.7ex\hbox{E}\kern-.125emX}}
\begin{document}

\title{Radiant: Large-scale 3D Gaussian Rendering based on Hierarchical Framework}

\author{
	\IEEEauthorblockN{
		Haosong Peng\IEEEauthorrefmark{1}\IEEEauthorrefmark{5}, 
		Tianyu Qi\IEEEauthorrefmark{2}\IEEEauthorrefmark{5},
		Yufeng Zhan\IEEEauthorrefmark{1}\IEEEauthorrefmark{6}, 
		Hao Li\IEEEauthorrefmark{3}, 
		Yalun Dai\IEEEauthorrefmark{4},
		Yuanqing Xia\IEEEauthorrefmark{1} 
	} 
	\IEEEauthorblockA{\IEEEauthorrefmark{1}School of Automation, Beijing Institute of Technology, Beijing, China}
	\IEEEauthorblockA{\IEEEauthorrefmark{2}School of Cyber Science and Technology, Sun Yat-sen University, Shenzhen, China}
    \IEEEauthorblockA{\IEEEauthorrefmark{3}Brain and Artificial Intelligence Lab, Northwestern Polytechnical University, Xi'an, China}
    \IEEEauthorblockA{\IEEEauthorrefmark{4}Nanyang Technological University, Singapore}
	\IEEEauthorblockA{livion@bit.edu.cn, qity9@mail2.sysu.edu.cn, yu-feng.zhan@bit.edu.cn, \\
	lifugan\_10027@outlook.com, daiy0018@e.ntu.edu.sg, xia\_yuanqing@bit.edu.cn}
}

\maketitle
\renewcommand{\thefootnote}{}
\footnotetext{\IEEEauthorrefmark{5}Equal Contribution.}
\footnotetext{\IEEEauthorrefmark{6}Corresponding author: yu-feng.zhan@bit.edu.cn}

\begin{abstract}
With the advancement of computer vision, the recently emerged 3D Gaussian Splatting (3DGS) has increasingly become a popular scene reconstruction algorithm due to its outstanding performance. 
Distributed 3DGS can efficiently utilize edge devices to directly train on the collected images, thereby offloading computational demands and enhancing efficiency. 
However, traditional distributed frameworks often overlook computational and communication challenges in real-world environments, hindering large-scale deployment and potentially posing privacy risks.
In this paper, we propose Radiant, a hierarchical 3DGS algorithm designed for large-scale scene reconstruction that considers system heterogeneity, enhancing the model performance and training efficiency.
Via extensive empirical study, we find that it is crucial to partition the regions for each edge appropriately and allocate varying camera positions to each device for image collection and training.
The core of Radiant is partitioning regions based on heterogeneous environment information and allocating workloads to each device accordingly. 
Furthermore, we provide a 3DGS model aggregation algorithm that enhances the quality and ensures the continuity of models' boundaries. 
Finally, we develop a testbed, and experiments demonstrate that Radiant improved reconstruction quality by up to 25.7\% and reduced up to 79.6\% end-to-end latency.
\end{abstract}


\section{Introduction}
3D Gaussian Splatting (3DGS) represents the latest advancement in the field of scene reconstruction of computer vision~\cite{kerbl20233d,wu2024recent}. 
Its ability to reconstruct real-world 3D structures directly from 2D images has led to widespread applications in virtual reality (AR/VR)~\cite{li2024vdg}, search-and-rescue~\cite{zhang2024darkgs}, and autonomous driving systems~\cite{sun2020scalability,li2024ggrt}, all aiming to achieve superior performance.
However, 3DGS-based methods pose challenges as they place a heavy load on the server and require extensive data storage and long completion time, especially when the scene is growing on a scale beyond the city scale. 

\begin{figure}[!t]
\begin{center}
\includegraphics[width=\linewidth]{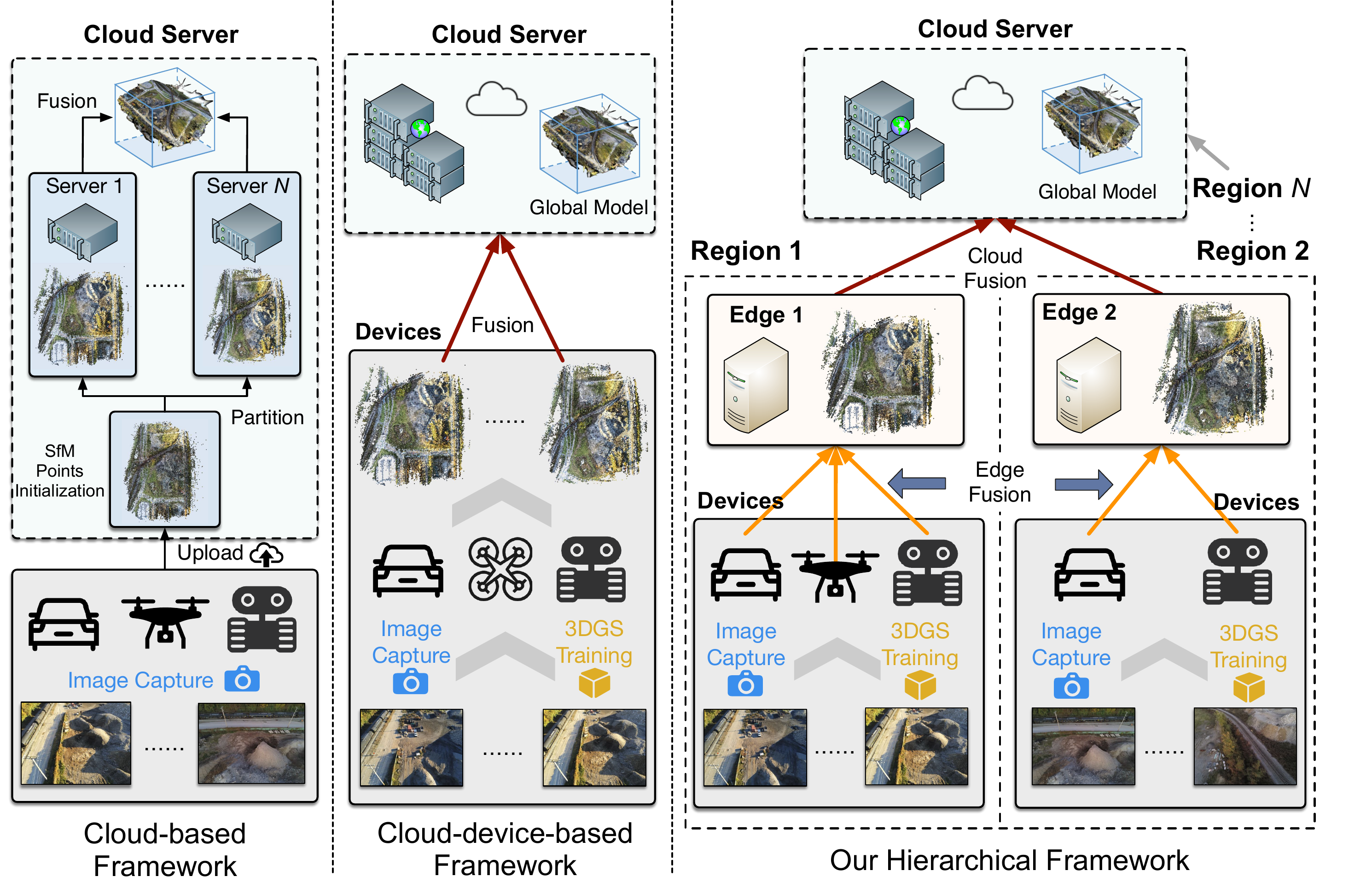}
\caption{\textbf{Left}: Cloud-based 3DGS training. \textbf{Middle:} Cloud-device-based 3DGS training. \textbf{Right:} Our hierarchical framework for 3DGS training.}
\label{fig:intro}
\end{center}
\end{figure}

%
Currently, many distributed 3DGS algorithms~\cite{lin2024vastgaussian,liu2024citygaussian,yuchen2024dogaussian} have been developed for large-scale scenarios.
A straightforward approach is cloud-based multi-GPU distributed training~\cite{lin2024vastgaussian}, as shown in the left part of Fig.~\ref{fig:intro}, which has shown excellent results but requires substantial resources. 
Take VastGaussian~\cite{lin2024vastgaussian} as an example, it requires two hours of training on 8 $\times$ Tesla V100 GPU and 2.6 GB storage for an area of only $500\times250m^2$.
Moreover, to obtain the initial model (i.e., the 3D Gaussian points) for the scene, all the raw images captured by the device must be uploaded to the cloud for structure-from-motion~\cite{snavely2006photo} (SfM),  not only increasing communication overhead but also severely infringing on device privacy.
Thanks to advancements in edge computing, cloud-device-based 3DGS has attracted attention \cite{suzuki2024fed3dgs,bao20243d}.
As shown in the middle part of Fig.~\ref{fig:intro}, multiple devices simultaneously capture images in different regions. 
These images undergo local 3DGS training, and the models are then uploaded to the cloud for fusion, enabling comprehensive scene reconstruction tasks.
However, cloud-device-based 3DGS algorithms do not address deployment issues in real-world scene reconstruction. 
The system's computational power and communication heterogeneity can cause stragglers.
As the scale of the scene increases, the uploading of models by numerous devices can lead to network congestion, and the fusion of models in the cloud incurs significant overhead. 
Additionally, the cloud can access the precise location privacy of the devices~\cite{zhao2023vector}.
Therefore, a more practical and efficient framework is urgently needed especially in large-scale scene reconstruction.

Recently, the cloud-edge-device framework has been proposed to address deployment issues in large-scale systems~\cite{liu2020client,wang2021resource,cui2022optimizing}. 
%
%
When this framework is applied to distributed 3DGS, as shown in the right part of Fig.~\ref{fig:intro}, each region can be effectively partitioned and trained. 
Specifically, the devices capture their images and independently train the models, which are then uploaded to the edge for preliminary fusion before being further fused in the cloud.
In this way, the computational loads are ingeniously reallocated from the cloud to the edge, which can significantly mitigate the burden in the cloud.
Besides, the images collected and the model trained by devices in different regions do not require sharing or direct uploading to the cloud, thus preserving data and location privacy.
%
Moreover, this framework effectively enhances the scalability of the system.
For new scenes, they can be developed as a new area by adding edges and devices to obtain a global model, without impacting the already trained regions.

Although combining the hierarchical framework with 3DGS can yield excellent results, the following challenges have to be addressed. 
In the real-world system, the pronounced heterogeneity can cause a severe straggler effect, ultimately affecting reconstructing efficiency~\cite{wang2021resource}. 
For example, the varying number of images on each device leads to differences in model sizes, resulting in heterogeneous communication time for fusion. 
Even with the same number of images, training time differs across devices with varying computational power. 
On the other hand, the geographical locations of the edges and the camera positions are often constrained in real-world environments.
Therefore, it is essential to consider how to partition regions geometrically for each edge and allocate workloads (i.e. camera positions) appropriately based on the heterogeneity of each device.
Furthermore, we find that traditional merging methods cause obvious visual disruptions at the boundaries after model fusion. 
This impact worsens with the increase in the number of models merged.
Developing an fusion scheme within the cloud-edge-device framework to maximize model effectiveness while addressing these issues remains a challenge.

In this work, we propose a hierarchical 3D Gaussian rendering framework for large scene reconstruction in a heterogeneous cloud-edge-device system, which jointly considers high efficiency, scalability, and privacy.
We introduce a new 3DGS model aggregation algorithm within the hierarchical framework, which effectively addresses the shortcomings of traditional merging algorithms.
On the other hand, we find that workload partitioning is crucial for the efficiency of modeling tasks in heterogeneous systems.
Different from other distributed 3DGS algorithms, this algorithm fully considers the computational and communication heterogeneity within a three-layer architecture.
Specifically, we propose an Adaptive Region Planning (ARP) algorithm based on geographical and resource characteristics to partition regions for edges, as well as a Resource-aware Task Partitioning (RTP) algorithm to allocate workloads for devices under each edge. 
We develop a prototype of Radiant and deploy it on a resource-heterogeneous testbed. 
Experiments on two real-world large-scale scene datasets show that Radiant can achieve better reconstruction performance and lower latency compared to the state-of-the-art. 
Our contributions are as follows:
\begin{itemize}
    \item To the best of our knowledge, we are the first to introduce the cloud-edge-device framework to large-scale scene reconstruction, and we design a model aggregation scheme tailored for this framework that maximizes model performance. 
    \item We develop an optimization algorithm to allocate workloads based on the real-world distribution of edges and camera positions, as well as the system heterogeneity, to minimize the end-to-end latency.
    \item We develop a prototype of Radiant and design a testbed, with experiments indicating that Radiant can reduce end-to-end latency by up to 79.6\% while achieving up to 25.7\% better quality compared to the state-of-the-art.
\end{itemize}

The rest of the paper is organized as follows: Section~\ref{sec2} shows the background and motivation of this paper. 
Section~\ref{sec3} provides a detailed design of Radiant.
We conduct extensive experiments in Section~\ref{sec5}, and the related work is reviewed in Section~\ref{sec6}. 
Finally, we conclude this paper in Section~\ref{sec7}.

\section{Background and Motivation}\label{sec2}

In this section, we first briefly review the background of 3D Gaussian Splatting.
Then we conduct an experimental study to motivate the design of Radiant.

\subsection{3D Gaussian Splatting}
Our scene reconstruction is based on 3DGS~\cite{kerbl20233d} technology, where the geometry and appearance of the scene are rendered using a set of 3D Gaussian. 
%
The cameras that capture images can be written as $\mathbf{C} = \{(\mathcal{C}_j, I_j)\}$, where $\mathcal{C}_j$ is the parameters of $j$-th camera including the position, intrinsic and extrinsic matrices, and $I_j$ is the corresponding image.
Let $\mathbf{G} = \{G_i\}$ be a set of 3D Gaussians.
Each 3D Gaussian $G_i = (\mathbf{x}_i, \mathbf{\Sigma}_i, \mathbf{f}_i, \alpha_i)$ is characterized by its position $\mathbf{x}_i=(x_i, y_i, z_i)$, covariance matrix $\mathbf{\Sigma}_i\in\mathbb{R}^{3\times3}$, opacity $\alpha_i\in\mathbb{R}$, and spherical harmonic coefficients $\mathbf{f}_i\in\mathbb{R}^l$ for view-dependent colors, where $l$ is related to the degree of the spherical harmonics. 
The initial position of 3D  Gaussian is initialized using SfM~\cite{snavely2006photo}.
3DGS renders the novel view image $I^r_j = \mathcal{R}(\mathcal{C}_j,\mathbf{G})$ by a differentiable rasterizer $\mathcal{R}$.

Specifically, 3D Gaussians are projected into the 2D image space for rendering.
The projected 2D covariance matrix $\mathbf{\Sigma}^\prime$ is
\begin{equation}
\boldsymbol{\Sigma}^{\prime}=\boldsymbol{J} \boldsymbol{W} \boldsymbol{\Sigma} \boldsymbol{W}^{\top} \boldsymbol{J}^{\top},
\end{equation}
where $\mathbf{W}\in\mathbb{R}^{3\times3}$ is the viewing transformation matrix and $J\in\mathbb{R}^{3\times3}$ is the Jacobian of the affine approximation of the projective transformation~\cite{zwicker2002ewa}.
Next, given the pixel $\mathbf{p}$ and its position, we get a list of Gaussians $\mathcal{N}$ sortd by their distance to the pixel.
Then, the color $C$ of each pixel is computed using $\alpha$-blending, as
\begin{equation}
C=\sum_{i \in \mathcal{N}} c_i \alpha_i^{\prime} \prod_{k=1}^{i-1}\left(1-\alpha_k^{\prime}\right),
\end{equation}
where $c_i$ is the learnable color computed from the spherical harmonics with $f_i$, and the final opacity $\alpha_i^\prime$ is the multiplication result of the learnable opacity $\alpha_i$ and the Gaussian, as
\begin{equation}
\alpha_i^{\prime}=\alpha_i \times \exp \left(-\frac{1}{2}\left(\mathbf{p}^{\prime}-\mathbf{x}_i^{\prime}\right)^{\top} \mathbf{\Sigma}_i^{\prime-1}\left(\mathbf{p}^{\prime}-\mathbf{x}_i^{\prime}\right)\right),
\end{equation}
where $\mathbf{p}^\prime$ and $\mathbf{x}^\prime$ are coordinates in the projected space.

The properties of 3D Gaussians are optimized with respect to the loss function between $I^r_j$ and the ground truth image $I_j$, as
\begin{equation}
\mathcal{L}=(1-\lambda) \mathcal{L}_1\left({I}_i^r, {I}_j\right)+\lambda \mathcal{L}_{\text{D-SSIM}}\left(I_j^r, I_j\right),
\end{equation}
where $\lambda$ is a hyper-parameter, and $\mathcal{L}_{\text{D-SSIM}}$ denotes the D-SSIM loss.
Finally, a set of trained 3D Gaussians $\mathbf{G}$ is referred to as a model.

\subsection{The Influence of Heterogeneity in 3DGS Training}
\begin{figure}[!t]
    \centering
    \begin{subfigure}{0.24\textwidth}
        \centering
        \includegraphics[width=1\linewidth]{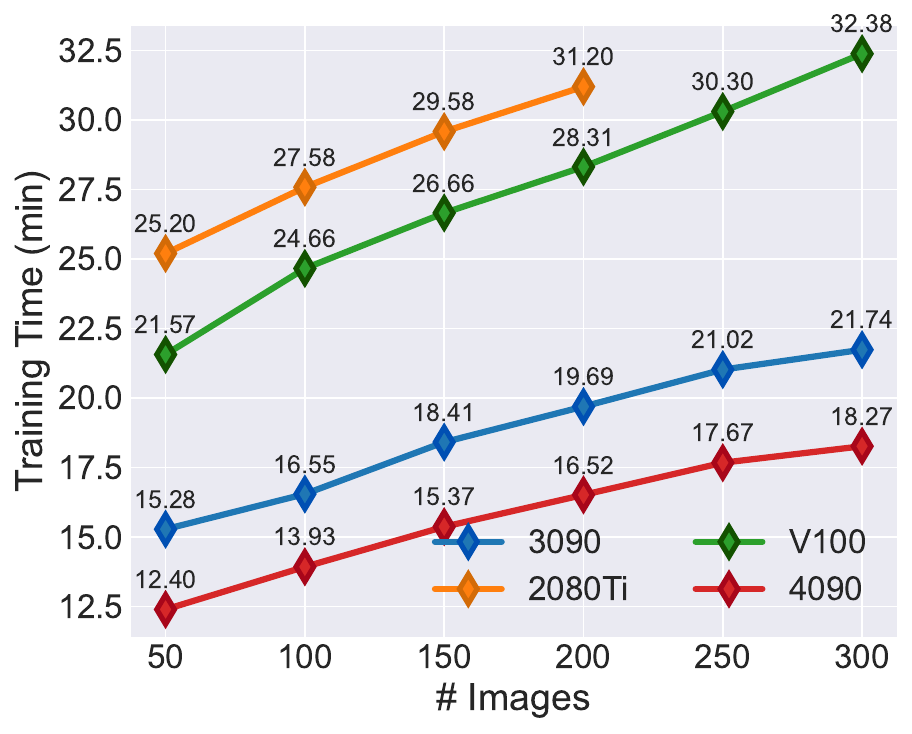}
        \caption{Training time v.s. \# Images}
        \label{fig2:sub1}
    \end{subfigure}
      \begin{subfigure}{0.24\textwidth}
        \centering
        \includegraphics[width=1\linewidth]{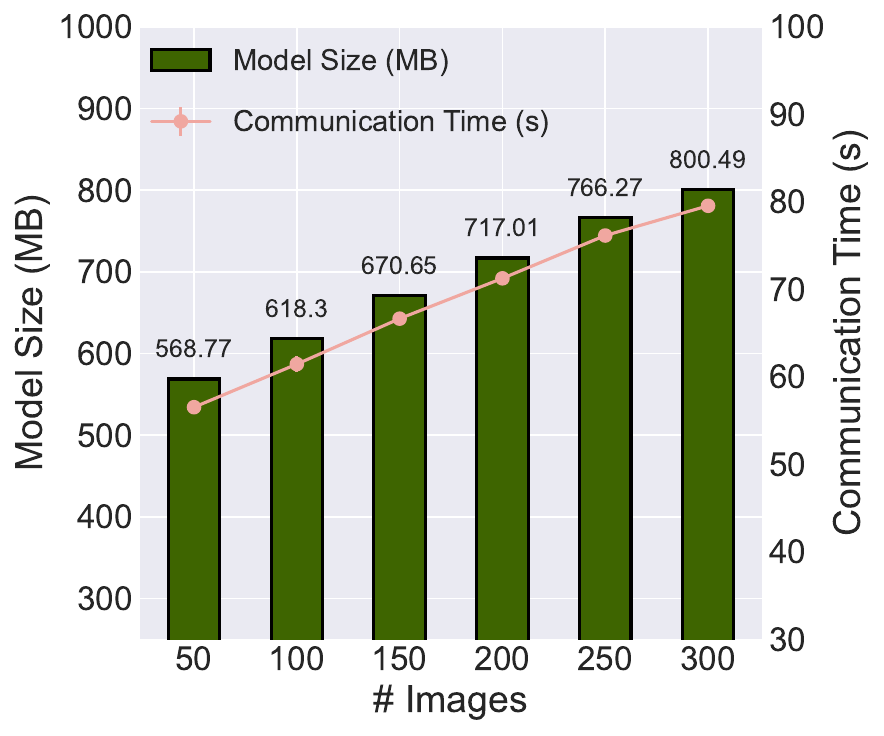}
        \caption{Model size and comm. time.}
        \label{fig2:sub2}
    \end{subfigure}
    \caption{Heterogeneity when training across different scene scales on various devices.}
    \label{motivation2}
\end{figure}

To explore the heterogeneity of 3DGS local device training in the real world, we conduct multiple experiments to verify the impact of varying image quantities.

\textbf{Device training time.}
To validate the influence on local model training time, we use four different devices with varying computational power: NVIDIA 4090, 3090, 2080Ti, and V100 to run the 3DGS training with varying quantities of images.
Fig.~\ref{fig2:sub1} illustrates training time across various scene scales with different devices. 
Generally, as the number of images increases, the scene construction of the model becomes more complex, leading to longer overall training time. 
%
%
It can be observed that device training time significantly vary across devices with different computational powers.

\textbf{Model size and communication time.}
We use the devices above to measure the model size and communication time under 5G bandwidth~\cite{ghoshal2022depth} after local training, as shown in Fig.~\ref{fig2:sub2}.
Since more images result in more 3D Gaussian being generated, leading to larger model sizes, there are still significant communication time differences even under 5G conditions, causing some stragglers.

The results show that the varying quantities of images on devices significantly impact local training time. 
This variation also results in models of different sizes, which subsequently affects the communication time to the edge and cloud for fusion.
Therefore, an effective workload partitioning algorithm is needed to ensure synchronous training across devices with various computational powers.

\subsection{The Influence of Geographical Region Partitioning}

In real-world scenarios, we need to assign nearby camera positions for image collection to each edge, thus forming a region. 
Devices within each region are designated to collect corresponding images.
Due to the varying number and computational power of devices in each region associated with an edge, different regions result in heterogeneous completion times.

\begin{figure}[!t]
    \centering
    \begin{subfigure}{0.16\textwidth}
        \centering
        \includegraphics[width=1\linewidth]{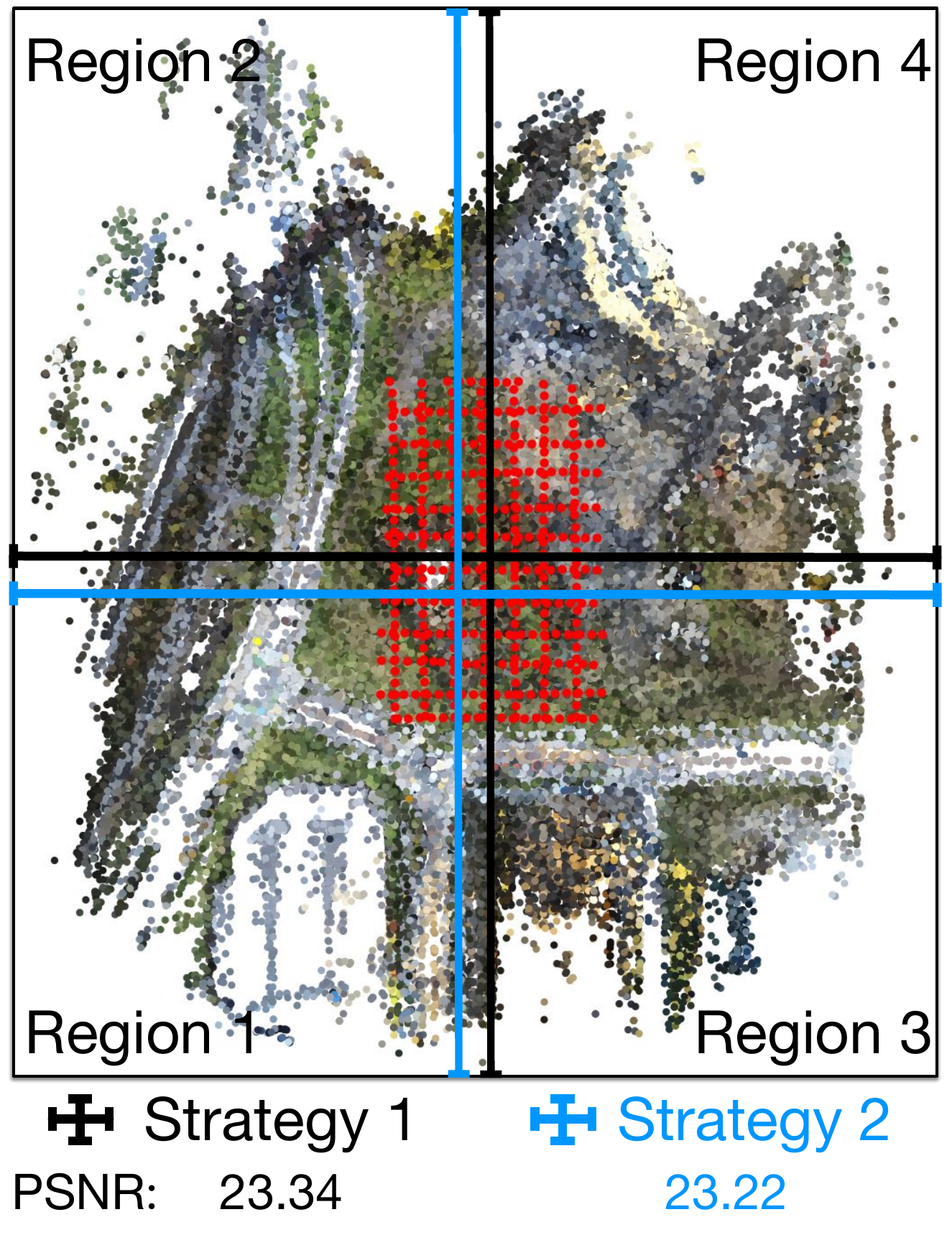}
        \caption{Partitioning strategy}
        \label{fig3:sub1}
    \end{subfigure}
      \begin{subfigure}{0.32\textwidth}
        \centering
        \includegraphics[width=1\linewidth]{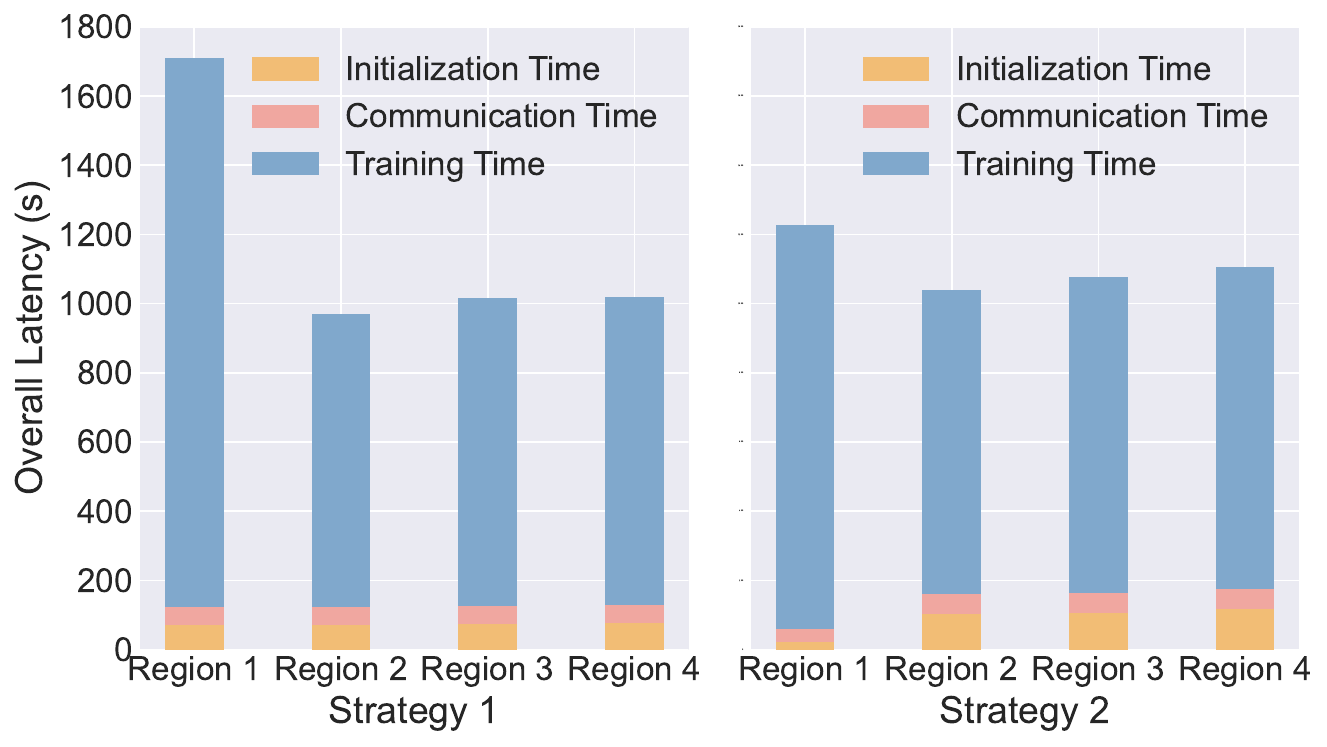}
        \caption{The overall latency of each region when equally partitioning and balanced partitioning. }
        \label{fig3:sub2}
    \end{subfigure}
    \caption{The overall latency breakdown of each device in two partitioning strategies.}
    \label{motivation3}
\end{figure}

To validate the importance of region partitioning, we test the overall latency using different partitioning strategies in the same scenario.
As shown in Fig.~\ref{fig3:sub1}, we deploy the low computational power devices in Region 1, while the high computational power devices were deployed in the other three regions.
We establish two partitioning strategies for the four regions. 
Partitioning strategy 1 represents evenly distributing camera positions across four regions, while strategy 2 reduces the number of cameras in Region 1 and offloads them to other regions.
The overall latency is illustrated in Fig.~\ref{fig2:sub2}, where it can be observed that Region 1 takes significantly longer.
On the contrary, strategy 2 has a 28\% latency reduction compared to the even distribution.
Nevertheless, the impact of the two partitioning strategies on model quality (i.e., PSNR) is very minimal.

The above experiments motivate that a reasonable geographical partition of regions can mitigate heterogeneity among edges. 
However, designing an optimal and refined partitioning algorithm for complex systems remains a challenge that needs to be addressed.

\subsection{The Influence of Different Fusion Scheme}
\begin{figure}[!t]
\begin{center}
\includegraphics[width=\linewidth]{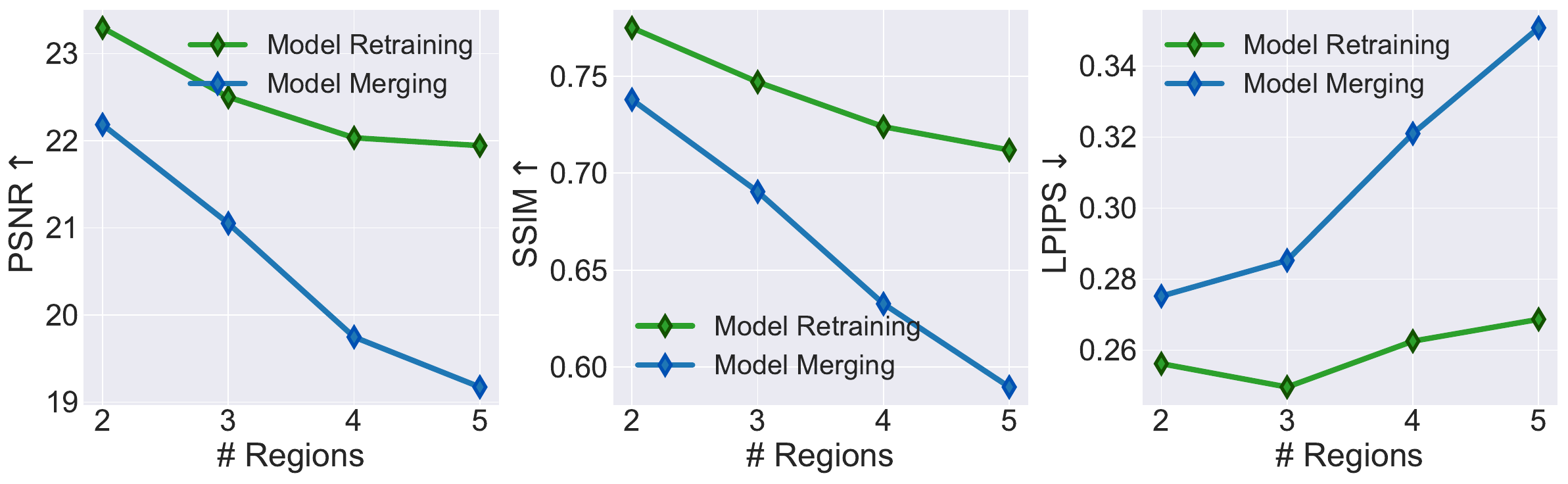}
\caption{The performance ($\uparrow$ PSNR, $\uparrow$ SSIM, $\downarrow$ LPIPS) of using model merge and model retrain algorithms in the same area.}
\label{motivation4}
\end{center}
\end{figure}

\begin{figure*}[!t]
    \centering
    \includegraphics[width=1\linewidth]{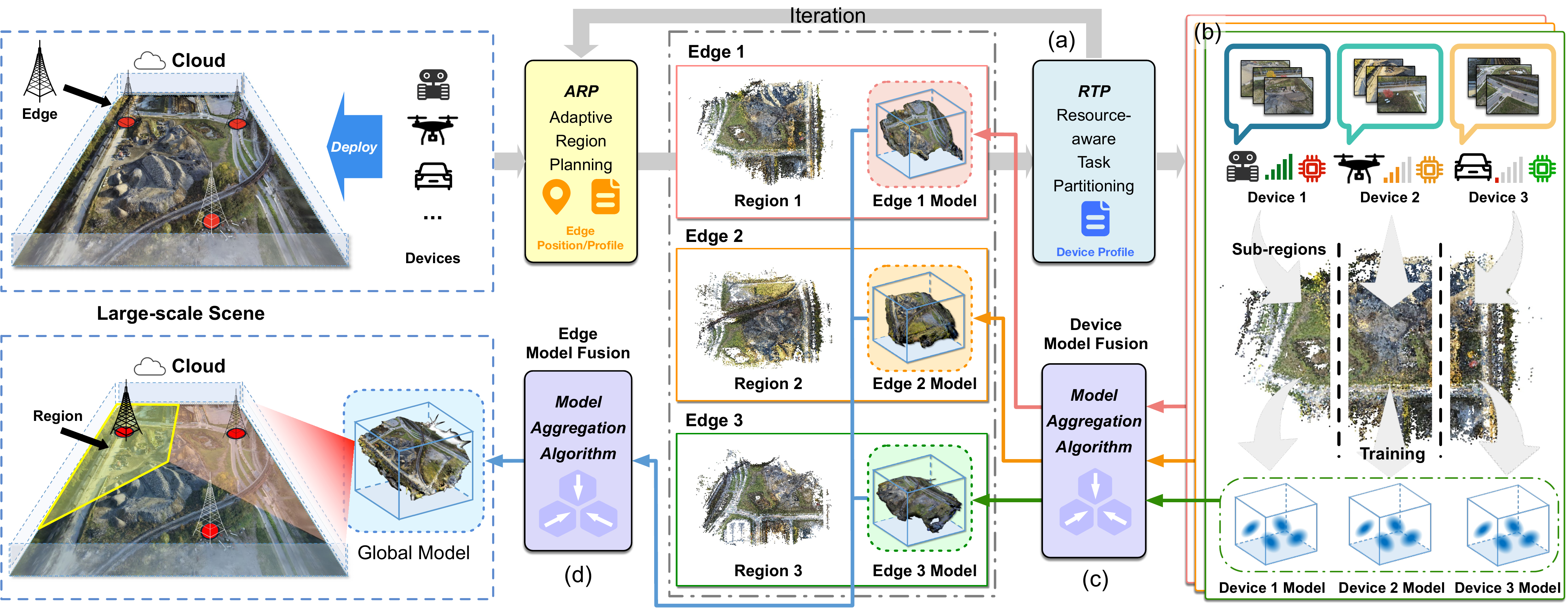}
    \caption{The Overview of Radiant. (a) The Adaptive Region Planning algorithm iterates together with the Resource-aware Task Partitioning algorithm. (b) Device training. (c) Device model aggregation. (d) Edge model aggregation.}
    \label{overview}
\end{figure*}
In distributed 3DGS systems, the common model merging algorithm~\cite{lin2024vastgaussian,yuchen2024dogaussian} directly cuts the model boundary and concatenates them together.
However, we observed that the merging method results in visual discontinuities at the boundaries (Fig.~\ref{fig:visual} visualizes this phenomenon).
To conduct a quantitative study, we select a scene and divide it into multiple regions for training.
We employ two fusion algorithms, a traditional merging method and a novel approach involving the upload of ground truth images followed by minimal retraining epochs of the entire model.
As shown in Fig.~\ref{motivation4}, we evaluate three quality metrics of the global fused model. 
As the number of regions increases, the performance of the merging method deteriorates, while retraining method results in negligible performance loss. 

The above results indicate that the merging method performs poorly due to boundary discontinuity, whereas retraining effectively mitigates this issue. 
However, retraining requires access to the ground truth images, which compromises the privacy of devices. 
Therefore, it is crucial to develop a fusion algorithm that utilizes retraining while preserving privacy.

\section{Radiant Design}\label{sec3}

In this section, we first present the overview of the Radiant framework. 
Then, the workload partitioning problems and algorithms are introduced, followed by the model aggregation algorithm.

\subsection{Overview}

Radiant employs a cloud-edge-device hierarchical architecture to achieve efficient scene reconstruction. 
As illustrated in Fig.~\ref{overview}, consider a large-scale scene with a set of camera positions where images will be collected.
This scene includes base stations at different positions, each serving as an edge center. 
Within each edge, there are multiple devices with various computational powers.
All the edges and devices should collaborate to reconstruct the entire scene.

The workflow of Radiant is introduced as follows:
(a) The large scene has to be divided into multiple regions, with its corresponding camera positions allocated to the edge.
The edge then distributes these camera positions among various devices for image collection and training.
%
%
(b) Then, all the devices independently collect their images and initialize the 3D Gaussian point by SfM, followed by training their local model.
(c) After all devices in the same edge complete their training, the models are uploaded to the edge for aggregation.
(d) All edge models are then uploaded to the cloud for the final aggregation, finally producing a global model of the scene.

%
Our design adheres to the following objectives:
(1) \textbf{Efficiency}. We allocate the workload across each device appropriately to minimize completion time while enhancing performance in the large-scale system.
(2) \textbf{Scalability}. Our framework allows new scenes to be added at any time, and once trained, these can be integrated with the models of previous scenes.
(3) \textbf{Privacy}. Images captured by devices are not uploaded to the edge or cloud, and there is no data exchange between devices. 
Additionally, the cloud does not have access to the location information of each device.
%

Specifically, to balance the workload on each edge, we propose the ARP algorithm to allocate the number and positions of cameras based on each region's computational power and edge's bandwidth (Sec.~\ref{ARP}). 
For device workload partitioning, we introduce the RTP Algorithm, which allocates the camera positions to each device based on computational power to mitigate straggler effects (Sec.~\ref{RTP}). 
Then, all the devices can conduct 3DGS training in parallel.
Next, we introduce a model aggregation algorithm designed to achieve high quality while ensuring rapid execution and privacy.
Finally, all the edge models are uploaded to the cloud and fused to form a global model (Sec.~\ref{dma}).

\subsection{Adaptive Region Planning Algorithm}\label{ARP}
Suppose there are $M$ edges $\mathcal{M} = \{1,2,\ldots\,M\}$ located at position $\mathcal{P}=\{P^1,\ldots,P^M\}$.
Edge $m$ has a set of devices $\mathcal{N}^m = \{1,2,\ldots,N^m\}$.
The communication bandwidth between each edge \( m \) and the cloud is denoted as \( B^m \), and the device bandwidth is $B^n$.
Therefore, the task completion time $T^m$ for each edge $m$ includes the training time $T^m_{train}$, the aggregation time $T^m_{aggre}$ and the communication time between edge and cloud $T^m_{com}$, which is 
\begin{equation}\label{eq5}
T^m = T_{train}^m + T^m_{aggre} + T_{com}^m.
\end{equation}  
$T_{com}^m$ is calculated as the edge model size $Z^m$ divided by the average bandwidth, defined as
\begin{equation}\label{tcom}
T^m_{com} = \frac{Z^m}{B^m},
\end{equation}
where \( Z^m \) can be approximated as the sum of the size of device models, i.e., $Z^m\approx\sum_{n=1}^{N^m}Z^n$.
The aggregation time $T^m_{aggre}$ is the time taken for all device models to be fused on the edge. 
In our method, this is essentially equivalent to model retraining, therefore, it is proportional to the number of images captured under each region.
The aggregation time of edge $m$ is calculated as:
\begin{equation}\label{eqagg}
T^n_{aggre} = \mathcal{F}_m(|\mathbf{C}^m|),
\end{equation}
where $|\mathbf{C}^m|$ is the amount of camera positions that assigned to edge $m$ and $\mathcal{F}_m$ is a function that can be fitted.
Within each edge, all devices can start training simultaneously, so the total completion time is determined by the device that takes the longest to complete its training:
\begin{equation}\label{ttrain}
T^m_{train} = \max_{n=1}^{N^m}\{T^n\}.
\end{equation}
The task completion time \( T^n \) of device $n$ is related to the number of camera positions allocated to it and its computational power, which can be calculated using the RTP algorithm introduced in Sec.~\ref{RTP}. 
In summary, to minimize the total completion time of the modeling task, we need to minimize the maximum completion time of any edge. 
This optimization problem can be formulated as
\begin{align}
&\min \max_{m=1}^{M} \left\{  T^m_{train} +T^m_{aggre} + T^m_{com} \right\}, \\
\text{s.t.} \quad & \bigcup_{m=1}^M \mathbf{C}^m = \mathbf{C},  \label{eq9} \\ 
& \mathbf{C}^i \cap \mathbf{C}^j = \emptyset,\quad \forall i, j \in \{1, 2, \ldots, M\}, \; i \neq j,  \label{eq10}\\
& \text{(\ref{tcom})}, \text{(\ref{eqagg})}, \text{(\ref{ttrain})} \notag. 
\end{align}
Eqn.~(\ref{eq9}) indicates that the region of each edge needs to cover all camera positions, while Eqn.~(\ref{eq10}) specifies that the regions must not overlap, thereby preserving the privacy of each edge.

The basic idea of the ARP is to achieve load balancing across all edges, which ensures that all edges complete their tasks in approximately the same amount of time.
Suppose we have the coordinates of all edge centers, the computational power of each device, and the positions of all images that need to be collected. 
The output is the final region of each edge and its camera position list.
The pseudo-code is shown in Algorithm~\ref{alg_ap}, which contains the following steps.

\begin{algorithm}[!t] 
	\caption{\label{alg_ap}Adaptive Region Planning Algorithm}
	\LinesNumbered 
	\KwIn{The positions of edges $\mathcal{P}=\{P^1,\ldots,P^M\}$, Camera position $\mathbf{C}$, distance $d$;} 
	\KwOut{Regions $ \mathbf{R}=\{r^1,\ldots,r^M\}$; Camera position lists $\mathbf{L}=\{l^1,\ldots,l^M\}$}
        $\mathbf{R}\leftarrow \text{Voronoi}(\mathcal{P})$ \#  Initialize the regions\;
        \For{$m = 1,2\ldots,M$}{
	$\mathbf{C}^m\leftarrow \text{Find}(\textbf{C},r^m)$ \# collect camera positions\;
        $T^m \leftarrow T_{train}^m + T^m_{aggre} + T_{com}^m$ \;
        }
        $\overline{T} \leftarrow \frac{1}{M}\sum_{m=1}^{M}T^m$ \;
        \While{$\max_m\{|T^m-\overline{T}|\}>T_{thres}$}
        {
        $m^*=\arg\max_m\{|T^m-\overline{T}|\}$\;
        $\mathbf{R}\leftarrow\text{Move\_boundary}(m^*,\mathbf{R},d)$\;
        \For{$m = 1,2\ldots,M$}{
        $\mathbf{C}^m\leftarrow \text{Find}(\textbf{C},r^m)$\;
        $T^m \leftarrow T_{train}^m + T^m_{aggre} +  T_{com}^m$\;
        }
        $\overline{T} \leftarrow \frac{1}{M}\sum_{m=1}^{M}T^m$ \;
        }
        \For{$m = 1,2\ldots,M$}{
        $l^m.\text{append}(\mathbf{C}^m)$\;
        }
\end{algorithm}

\begin{enumerate}
    \item \textbf{Initializing the regions.} We generate the initial division of $N$ regions (Line 1) following the rule of the Voronoi diagram~\cite{aurenhammer1991voronoi}.
    In the Voronoi diagram, the distance from all camera positions within each region to their respective edge center $\mathcal{P}$ is shorter than any other edge center.
    This partitioning ensures that devices can collect images in the nearest area, enhancing efficiency in the real systems.
    \item \textbf{Estimate the completion time.} The camera positions within each region are incorporated into the corresponding edge (Line 2).
    We estimate the completion time for all edges using Eqn.~(\ref{eq5}), then calculate their average time $\overline{T}$ (Lines 3-4).
    \item \textbf{Move the boundary of the outlier.} We select the edge with the greatest deviation from the average estimated time, determined by:
    \begin{equation}
    m^*=\arg\max_m\{|T^m-\overline{T}|\}.
    \end{equation}
    If $T^{m^*}$ of edge $m^*$ exceeds the average $\overline{T}$, the boundary is shrunk by a distance \(d\) in the direction of the boundary's normal; if it is below the average, the boundary is expanded by \(d\) in the normal direction (Line 7). 
    \item \textbf{Repeat until the condition is satisfied.}
    At each round, the completion time of each edge for the new regions can be re-estimated, and the process is repeated until the residual completion time of every edge is below a specified threshold (Lines 5-11).
\end{enumerate}

\begin{figure}[t]
    \centering
    \includegraphics[width=1\linewidth]{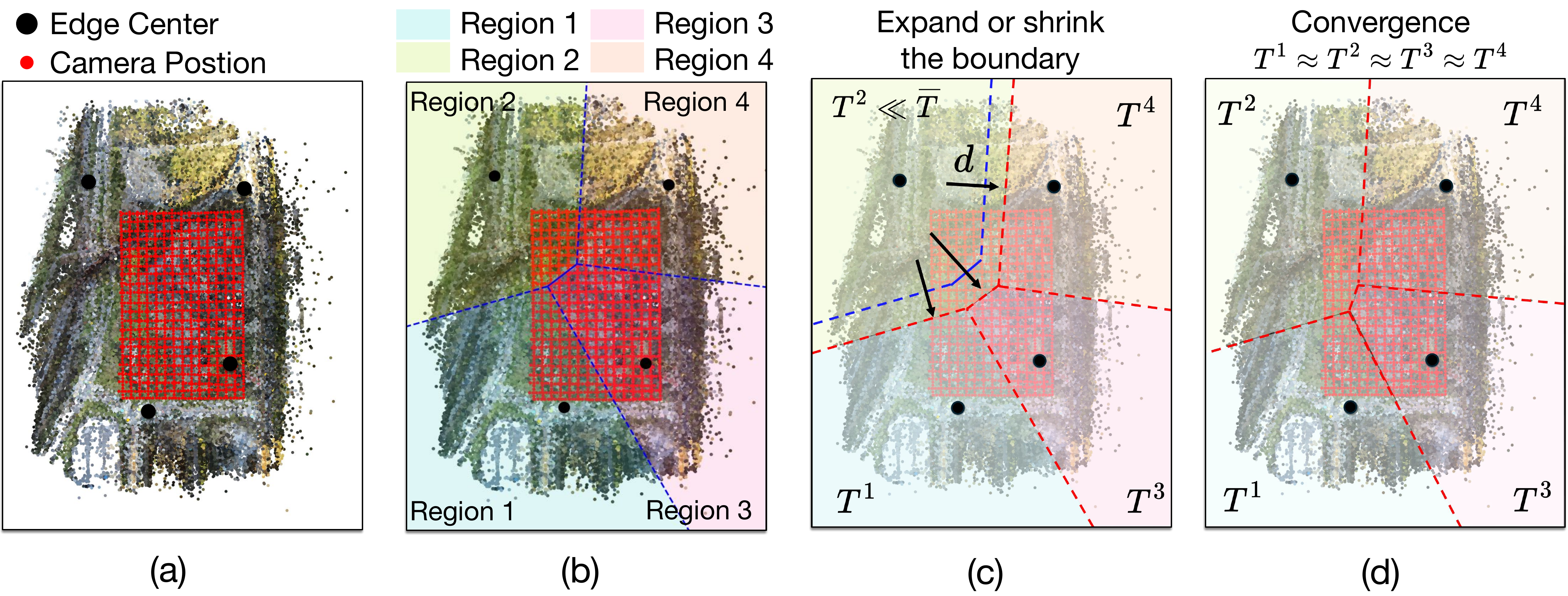}
    \caption{An example of Adaptive Region Planning algorithm. 
    (a) Original map of camera positions and edge centers. (b) Initialize the regions with a Voronoi diagram. (c) Iteratively adjusts the boundaries of the region to balance the load. (d) Converges to equilibrium.}
    \label{alg1}
\end{figure}

Fig.~\ref{alg1} illustrates an example of the ARP algorithm. 
The four edge centers, marked by black circles, are situated at distinct positions, while the red dots represent the camera position.
We initially partitioned these into four regions using a Voronoi diagram. 
However, this allocation results in a highly imbalanced workload on each edge because of the heterogeneity of devices. 
For example, region 2 is allocated very few cameras and consequently has a very fast completion time, which is identified as an outlier.
Therefore, we expand its boundary by \(d\) units and the completion time can be re-estimated.
We continue to iterate this process until the completion times of all four edges converge towards the same.
Finally, we record the positions and amount of the cameras assigned to each edge and the boundaries of their respective regions.
The ARP algorithm benefits from favorable initial conditions and can quickly converge through rapid iterations, efficiently allocating regions to edges without overlap within several seconds.

\subsection{Resource-aware Task Partitioning Algorithm}\label{RTP}
\begin{figure}[t]
    \centering
    \includegraphics[width=1\linewidth]{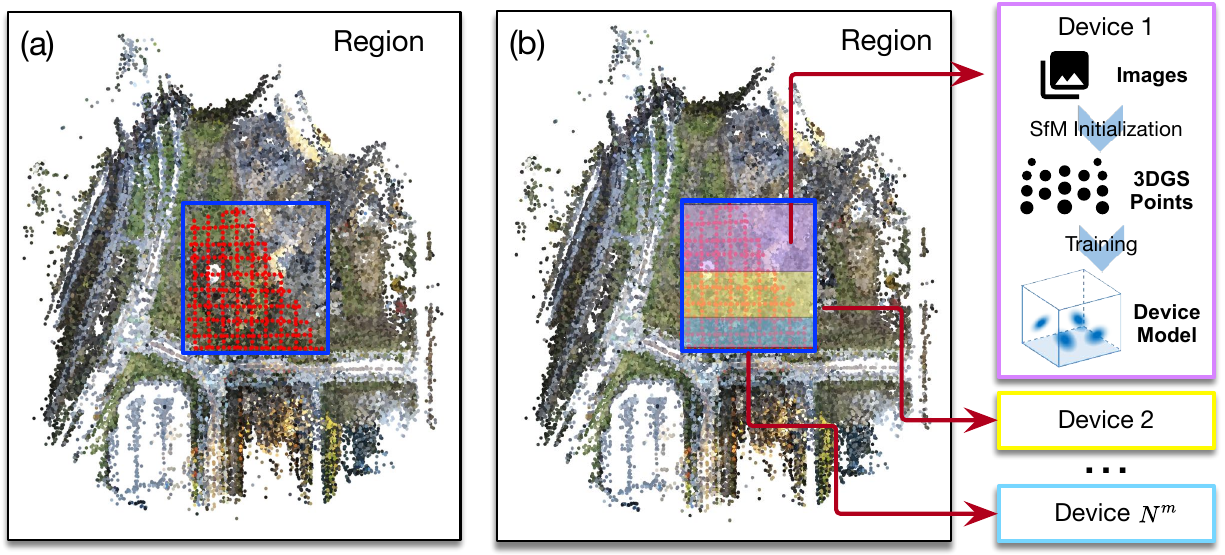}
    \caption{An example of Resource-aware Task Partitioning algorithm.} 
    %
    %
    \label{alg2}
\end{figure}

At each iteration of the ARP algorithm, we need to estimate the optimal completion time for each edge to complete its tasks. 
This problem can be modeled as a load-balancing assignment issue, with the objective of minimizing the longest completion time among all devices.
We already know the partitions of camera position quantities to each edge and need to allocate them to the resource-heterogeneous devices next.

For simplicity, we only consider the situation in edge $m$ as the problem is identical within each edge.
Specifically, the completion time for each device $n$ equals the sum of the initialization time, training time, and the communication time between the device and the edge, which is $T^n=T^n_{init}+T_{train}^n + T_{com}^n$.
The initialization time for 3D Gaussian is related to the image number used by the device, which can be profiled by
\begin{equation}\label{eq12}
T^n_{init} = \mathcal{G}(|\mathbf{C}^n|).
\end{equation}
The communication time is calculated as the device model size $Z^n$ divided by the average bandwidth, and the model size is related to the number of cameras, defined as:
\begin{equation}
Z^n = \mathcal{H}(|\mathbf{C}^n|),
\end{equation}
where function $\mathcal{H}(\cdot)$ can be profiled offline, as illustrated in  Fig.~\ref{fig2:sub2}.
Therefore, $T^n_{com}$ is calculated by 
\begin{equation}\label{tncom}
T^n_{com} = \frac{Z^n}{B^n}.
\end{equation}
Furthermore, we can determine the function between the training time and different cameras for devices with various computational powers by offline profiling as illustrated in Fig.~\ref{fig2:sub1}.
Hence, the training time is 
\begin{equation}\label{tntrain}
T^n_{train} = \mathcal{F}_n(|\mathbf{C}^n|),
\end{equation}
Overall, this problem can be formulated as:
\begin{align}
&\min \max_{n=1}^{N^m} \left\{ T^n_{init} + T^n_{train} + T^n_{com} \right\}, \label{eqrtp}\\
\text{s.t.} \quad  & \bigcup_{n=1}^{N^m} \mathbf{C}^n = \mathbf{C}^m, \\
& \mathbf{C}^i \cap \mathbf{C}^j = \emptyset,\quad \forall i, j \in \{1, 2, \ldots, N\}, \; i \neq j, \\
& |\mathbf{C}^n| \leq |\mathbf{C}^n_{max}|. \label{eq13} \\
& \text{(\ref{eq12})}, \text{(\ref{tncom})}, \text{(\ref{tntrain})} \notag
\end{align}
Eqn.~(\ref{eq13}) stipulates that the number of camera positions allocated to each device must not exceed the maximum capacity that its video memory can handle.
The optimal allocation will be achieved when the completion times of all devices are equal.
Consider that $\mathcal{G}$, $\mathcal{F}$, and $\mathcal{H}$ are known, the number of camera positions allocated to each device can be directly calculated. 
The result serves as the estimated values for $T^m_{train}$ from Eqn.~(\ref{ttrain}), which contributes to the iterations of the ARP algorithm.

%

%
Fig.~\ref{alg2} presents an example of the RTP algorithm.
Without loss of generality, we divide the camera positions into $N^m$ sub-regions along the longer axis of the rectangular area they occupy according to any device order.
%
%
The number of images captured by cameras for each sub-region is determined by the solution to problem~(\ref{eqrtp}). 
%
%
%
Finally, each device begins by initializing the Gaussian points using the captured images, followed by 3DGS training for reconstructing a specific sub-region.
The combined scope of these sub-regions is sufficient to cover the entire region of the edge.
%

Through the ARP and RTP algorithms, the reconstruction tasks for the entire scene are meticulously distributed to each device. 
This ensures that all devices can complete their respective tasks as efficiently as possible while mitigating the effects of stragglers.

\subsection{Model Aggregation Algorithm}\label{dma}

\begin{figure}[t]
    \centering
    \includegraphics[width=1\linewidth]{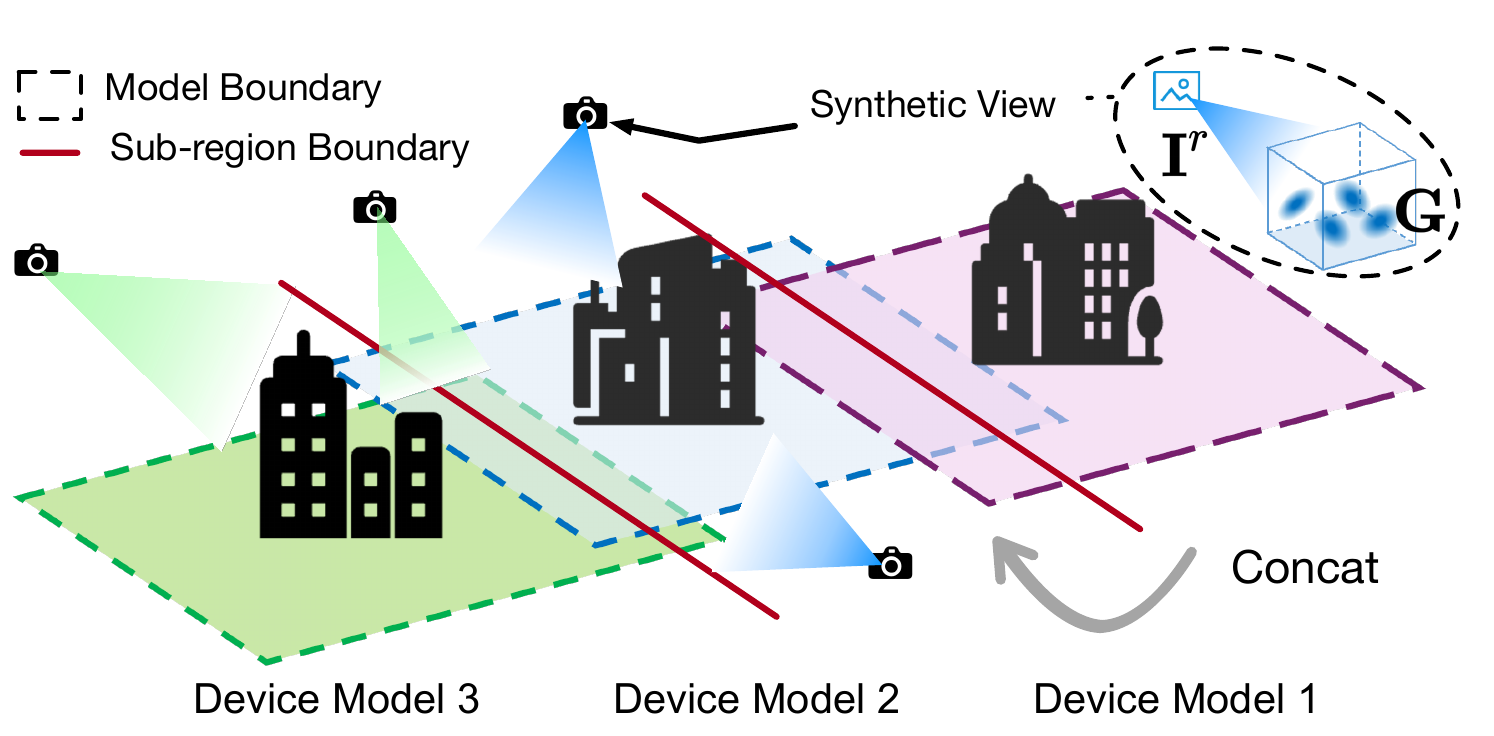}
    \caption{Model Aggregation. 
    We concatenate device models directly to form an edge model and retrain it using synthetic views.}
    \label{figdma}
\end{figure}

%
Traditional cutting and merging according to region boundaries does not yield an ideal reconstruction quality. 
%
%
To address this issue, we propose a model aggregation algorithm for the edge that seamlessly stitches together all device models. 
Specifically, each device model is trained using non-overlapping cameras $\mathbf{C}^n$. 
Synthetic views $\mathbf{I}^r$ can be rendered based on the perspectives of camera $\mathbf{C}^n$ after the devices upload their model to the edge, which can be expressed as $\mathbf{I}^r = \mathcal{R}(\mathcal{C}^n,\mathbf{G}^n)$, where $\mathbf{G}^n$ is the model of device $n$.
We concatenate all device models together and retrain the edge model for a few epochs using these synthetic views $\mathbf{I}^r$ as depicted in Fig.~\ref{figdma}. 
This approach can significantly enhance the quality of the model boundaries while eliminating the need to upload images to the edge, protecting the privacy of devices.

The aggregation of cloud models follows the same steps. 
In cases where cloud resources are insufficient for aggregation, we can still employ the merging method.
Specifically, according to the final region boundaries recorded in the ARP algorithm, we cut out the 3D Gaussians outside the boundary lines of each region from the edge models, and then directly concatenate them.

\section{Evaluation}\label{sec5}
We first describe the implementation of Radiant and experimental settings, followed by Radiant performance and its latency comparison.
Finally, we conduct ablation studies and visualization to prove the effectiveness of the Radiant.

\subsection{Implementation}\label{sec4}

Our Radiant framework is implemented with Python.
We use the HUAWEI Elastic Cloud Server (ECS)~\cite{huawei} with 8 vCPU and 16G memory as the cloud server.
In addition, we use the following servers with different GPUs as the heterogeneous device set.
\textbf{Instance \#1}: A desktop equipped with two Intel(R) Xeon(R) Silver 4310 CPUs, 128 GB of RAM, and two NVIDIA 4090 GPUs, each with 24 GB of VRAM.
\textbf{Instance \#2}: A desktop equipped with an Intel(R) Xeon(R) Silver 4310 CPU, 377 GB of RAM, and eight NVIDIA V100 GPUs, each with 32 GB of VRAM. 
\textbf{Instance \#3}: A desktop equipped with an Intel(R) Core(TM) i5-9400F CPU, 24 GB of RAM, and an NVIDIA 2080Ti GPU with 11 GB of VRAM.
\textbf{Instance \#4}: A desktop equipped with an Intel(R) Core(TM) i7-13700KF CPU, 64 GB of RAM, and an NVIDIA 3090Ti GPU with 24 GB of VRAM.
We use 5G bandwidth data in our system\cite{ghoshal2022depth}.

\subsection{Experiment Settings}
\subsubsection{Performance Metrics} The performance metrics we evaluate include end-to-end latency and global model quality. 
We assess the quality of scene reconstruction using three quantitative metrics: PSNR, SSIM~\cite{wang2004image} and VGG-based LPIPS~\cite{zhang2018unreasonable}. 
%

\subsubsection{Datasets} We use the Mill 19 dataset~\cite{turki2022mega} which consists of \textit{Rubble} and \textit{Building} scenes.
The \textit{Rubble} and \textit{Building} datasets consist of 1,678 and 1,940 images respectively with a resolution of 4608$\times$3456. 
The two scenes together cover a large-scale area of approximately 100,000 $m^2$.
We resize them to 1152$\times$864 for training and evaluation, following the same setting of~\cite{lin2024vastgaussian}. 

\subsubsection{Training Hyper-parameters} Each device is optimized for 20,000 iterations. 
The densification~\cite{kerbl3Dgaussians} starts at the 500-th iteration and ends at the 15,000-th iteration, with an interval of 100 iterations.
The number of epochs for retraining during model aggregation is set to 10.
The other settings are identical to those of 3DGS~\cite{kerbl3Dgaussians}.

\subsubsection{Baselines} We compare Radiant with centralized and distributed training methods.
We will also include the NeRF-based~\cite{mildenhall2021nerf} 3D reconstruction method in our comparison.
The centralized approach includes GP-NeRF~\cite{zhang2023efficient}, and 3DGS~\cite{kerbl3Dgaussians}, Switch-NeRF~\cite{zhenxing2022switch}, and the distributed method includes VastFaussian~\cite{lin2024vastgaussian}, Mega-NeRF~\cite{turki2022mega}, Drone-NeRF~\cite{jia2024drone}, FedNeRF~\cite{suzuki2023federated}, and Fed3DGS~\cite{suzuki2024fed3dgs}. 
Additionally, we present the privacy and scalability features of each method to ensure a fair comparison.

\subsection{Radiant Performance}
\begin{table}[t]
\centering
\caption{Composition of the heterogeneous systems.}
\label{system_transposed}
\begin{tabular}{@{}c|c|c@{}}
\toprule
 & \textbf{System 1} & \textbf{System 2} \\ \midrule
\textbf{Cloud} & HUAWEI Cloud ECS & HUAWEI Cloud ECS \\ \midrule
\textbf{Edge} & 1$\times$ 4090 & 1$\times$ 4090 \\ \midrule
\textbf{Region 1} & Instance \#1 & Instance \#1 \\
\textbf{Region 2} & Instance \#1 \& Instance \#4 & Instance \#1 \& Instance \#4 \\
\textbf{Region 3} & Instance \#2 \& Instance \#3 & Instance \#2 \& Instance \#3 \\
\textbf{Region 4} & - & Instance \#2 \& Instance \#4 \\ \bottomrule
\end{tabular}
\end{table}

\begin{table*}[!t]
\centering
\caption{Performance of Radiant and several baselines.}
\label{main results}
\begin{threeparttable}
\begin{tabular}{@{}cc|c|ccc|ccc|cc@{}}
\toprule
\multicolumn{2}{c|}{\multirow{2}{*}{\textbf{Architecture}}} &
  \multirow{2}{*}{\textbf{Method}} &
  \multicolumn{3}{c|}{\textbf{Rubble}} &
  \multicolumn{3}{c|}{\textbf{Building}} &
  \multirow{2}{*}{\textbf{Scalability}} &
  \multirow{2}{*}{\textbf{Privacy}} \\ \cmidrule(lr){4-9}
\multicolumn{2}{c|}{} &
   &
  \multicolumn{1}{c|}{\textbf{PSNR}$\uparrow$} &
  \multicolumn{1}{c|}{\textbf{SSIM}$\uparrow$} &
  \textbf{LPIPS} $\downarrow$ &
  \multicolumn{1}{c|}{\textbf{PSNR}$\uparrow$} &
  \multicolumn{1}{c|}{\textbf{SSIM}$\uparrow$} &
  \textbf{LPIPS} $\downarrow$ &
   &
   \\ \midrule
\multicolumn{2}{c|}{\multirow{3}{*}{\textbf{\begin{tabular}[c]{@{}c@{}}Centralized \\ Framework\end{tabular}}}} &
  \textbf{GP-NeRF~\cite{zhang2023efficient}} &
  \multicolumn{1}{c|}{24.08} &
  \multicolumn{1}{c|}{0.563} &
  0.497 &
  \multicolumn{1}{c|}{20.99} &
  \multicolumn{1}{c|}{0.565} &
  0.490 &
  \multicolumn{1}{c|}{\XSolidBrush} &
  \XSolidBrush \\
\multicolumn{2}{c|}{} &
  \textbf{3DGS\tnote{$\diamondsuit$} \cite{kerbl3Dgaussians}} &
  \multicolumn{1}{c|}{25.51} &
  \multicolumn{1}{c|}{0.725} &
  0.316 &
  \multicolumn{1}{c|}{22.53} &
  \multicolumn{1}{c|}{0.738} &
  0.214 &
  \multicolumn{1}{c|}{\XSolidBrush} &
  \XSolidBrush \\
\multicolumn{2}{c|}{} &
  \textbf{Switch-NeRF~\cite{zhenxing2022switch}} &
  \multicolumn{1}{c|}{24.31} &
  \multicolumn{1}{c|}{0.562} &
  0.496 &
  \multicolumn{1}{c|}{21.54} &
  \multicolumn{1}{c|}{0.579} &
  0.474 &
  \multicolumn{1}{c|}{\XSolidBrush} &
  \XSolidBrush \\ \midrule
\multicolumn{1}{c|}{\multirow{9}{*}{\textbf{\begin{tabular}[c]{@{}c@{}}Distributed  \\ Framework \end{tabular}}}} &
  \multirow{3}{*}{\textbf{\begin{tabular}[c]{@{}c@{}}Cloud-based \end{tabular}}} &
  \textbf{Drone-NeRF~\cite{jia2024drone}} &
  \multicolumn{1}{c|}{19.51} &
  \multicolumn{1}{c|}{0.528} &
  0.489 &
  \multicolumn{1}{c|}{18.46} &
  \multicolumn{1}{c|}{0.490} &
  0.469 &
  \multicolumn{1}{c|}{\Checkmark} &
  \XSolidBrush \\
\multicolumn{1}{c|}{} &
   &
  \textbf{Mega-NeRF~\cite{turki2022mega}} &
  \multicolumn{1}{c|}{24.06} &
  \multicolumn{1}{c|}{0.553} &
  0.516 &
  \multicolumn{1}{c|}{20.93} &
  \multicolumn{1}{c|}{0.547} &
  0.504 &
  \multicolumn{1}{c|}{\Checkmark} &
  \XSolidBrush \\
\multicolumn{1}{c|}{} &
   &
  \textbf{VastGaussian~\cite{lin2024vastgaussian}} &
  \multicolumn{1}{c|}{25.20} &
  \multicolumn{1}{c|}{0.742} &
  0.264 &
  \multicolumn{1}{c|}{21.80} &
  \multicolumn{1}{c|}{0.728} &
  0.225 &
  \multicolumn{1}{c|}{\XSolidBrush} &
  \XSolidBrush \\ \cmidrule(l){2-11} 
\multicolumn{1}{c|}{} &
  \multirow{2}{*}{\textbf{\begin{tabular}[c]{@{}c@{}}Cloud-device-based\end{tabular}}} &
  \textbf{FedNeRF~\cite{suzuki2023federated}} &
  \multicolumn{1}{c|}{20.12} &
  \multicolumn{1}{c|}{-} &
  - &
  \multicolumn{1}{c|}{17.51} &
  \multicolumn{1}{c|}{-} &
  - &
  \multicolumn{1}{c|}{\Checkmark} &
  \XSolidBrush \\
\multicolumn{1}{c|}{} &
   &
  \textbf{Fed3DGS~\cite{suzuki2024fed3dgs}} &
  \multicolumn{1}{c|}{20.62} &
  \multicolumn{1}{c|}{0.588} &
  0.437 &
  \multicolumn{1}{c|}{18.66} &
  \multicolumn{1}{c|}{0.602} &
  0.362 &
  \multicolumn{1}{c|}{\Checkmark} &
  \XSolidBrush \\ \cmidrule(l){2-11} 
\multicolumn{1}{c|}{} &
  \multirow{4}{*}{\textbf{\begin{tabular}[c]{@{}c@{}}Hierarchical \end{tabular}}} &
  \textbf{Even-system1\tnote{$\triangle$}} &
  \multicolumn{1}{c|}{23.93} &
  \multicolumn{1}{c|}{0.719} &
  0.324 &
  \multicolumn{1}{c|}{21.19} &
  \multicolumn{1}{c|}{0.682} &
  0.304 &
  \multicolumn{1}{c|}{\Checkmark} &
  \Checkmark \\
\multicolumn{1}{c|}{} &
   &
  \textbf{Even-system2\tnote{$\triangle$}} &
  \multicolumn{1}{c|}{24.08} &
  \multicolumn{1}{c|}{0.729} &
  0.310 &
  \multicolumn{1}{c|}{21.33} &
  \multicolumn{1}{c|}{0.698} &
  0.297 &
  \multicolumn{1}{c|}{\Checkmark} &
  \Checkmark \\
\multicolumn{1}{c|}{} &
   &
  \textbf{Radiant-system1} &
  \multicolumn{1}{c|}{24.44} &
  \multicolumn{1}{c|}{0.737} &
  0.305 &
  \multicolumn{1}{c|}{21.57} &
  \multicolumn{1}{c|}{0.716} &
  0.292 &
  \multicolumn{1}{c|}{\Checkmark} &
  \Checkmark \\
\multicolumn{1}{c|}{} &
   &
  \textbf{Radiant-system2} &
  \multicolumn{1}{c|}{\textbf{24.53}} &
  \multicolumn{1}{c|}{\textbf{0.740}} &
  \textbf{0.284} &
  \multicolumn{1}{c|}{\textbf{21.66}} &
  \multicolumn{1}{c|}{\textbf{0.723}} &
  \textbf{0.236} &
  \multicolumn{1}{c|}{\Checkmark} &
  \Checkmark \\ \bottomrule
\end{tabular}
 \begin{tablenotes}
        \footnotesize
        \item [$\diamondsuit$] modified 3DGS~\cite{kerbl3Dgaussians} so that it can be optimized on a single instance;  $^\triangle$ ``even" means to divide the images equally for each device.
  \end{tablenotes}
\end{threeparttable}
\end{table*}

To demonstrate the performance of Radiant in large heterogeneous systems, we set up two groups of heterogeneous systems to complete scene reconstruction tasks in the \textit{Rubble} and \textit{Building} scene, respectively.
Table~\ref{system_transposed} shows the composition of the two systems we used, the number of edges, and the configuration of edges. 
It is worth noting that each edge comprises several devices simulated by the \textbf{Instance} as mentioned in Sec~\ref{sec4}.
The locations of edge servers are randomly and uniformly distributed throughout the scene.

Table~\ref{main results} presents the performance of Radiant alongside several baseline methods. 
We categorize the methods into centralized and distributed frameworks, with the latter further divided into cloud-based, cloud-device-based, and hierarchical frameworks.
We also compare another Radiant version that evenly divides the workload to each device.
As a novel method based on the hierarchical framework, Radiant outperforms all cloud-device framework methods in terms of PSNR, SSIM, and LPIPS. 
Compared to centralized and cloud-based architecture methods, it achieves comparable performance.
However, the centralized method based on 3DGS requires collecting all images and initializing Gaussian points before training, thus failing to protect privacy. 
Similarly, it lacks scalability when new areas need to be planned and reconstructed.
In FedNeRF and Fed3DGS, devices directly upload their models to the cloud, revealing each device's location and thus failing to satisfy privacy requirements.
Our Radiant method achieves high-quality reconstruction whether the workload is evenly distributed or divided according to heterogeneous systems. 
Specifically, Radiant can improve reconstruction quality (PSNR) by up to 25.7\%.
Besides, it satisfies both privacy and scalability requirements.

\subsection{Latency Comparison}

\begin{figure}[!t]
\begin{center}
\includegraphics[width=\linewidth]{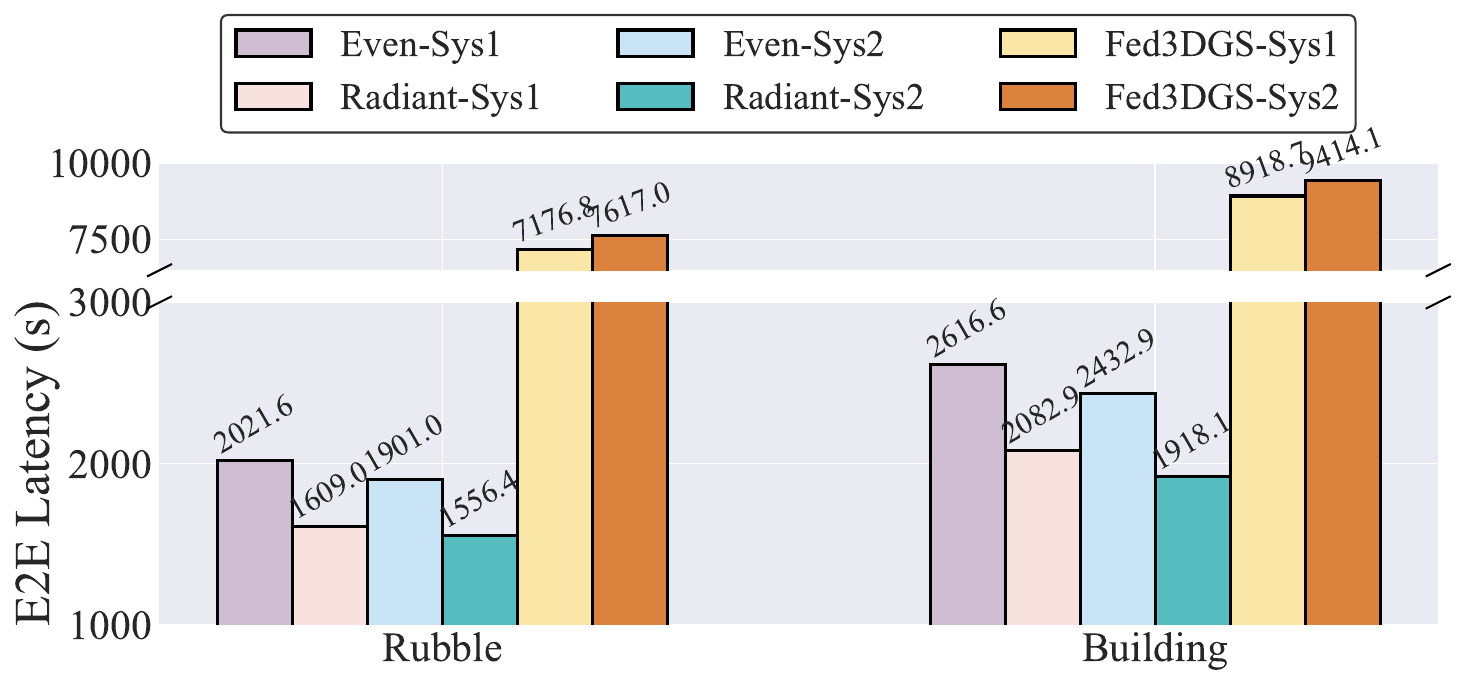}
\caption{Comparison of end-to-end latency. ``even" means to divide the images equally for each device.}
\label{fig:latency}
\end{center}
\end{figure}

We compare the latency of Radiant with Fed3DGS~\cite{suzuki2024fed3dgs} in Fig.~\ref{fig:latency}. 
We implement the experiments using the Fe3DGS under the same scale of System~1 and System~2. 
Methods based on the NeRF usually require more than 1 day~\cite{turki2022mega,lin2024vastgaussian} training, making them impractical for timely applications, and thus are excluded from our comparison. 
Specifically, Radiant is 19.27\% and 20.99\% faster than even distribution in the \textit{Rubble} and \textit{Building} scenarios, respectively, because even distribution slows down the completion time of the devices with the least computational resources.
All our Radiant versions are more than 70\% faster than Fed3DGS. 
For instance, our Radiant-system2 is 79.6\% faster than Fededgs on \textit{Building} dataset. 
This is due to Fed3DGS fusing all device modes sequentially on a single machine, which is highly time-consuming.

\begin{figure}[!t]
\begin{center}
\includegraphics[width=\linewidth]{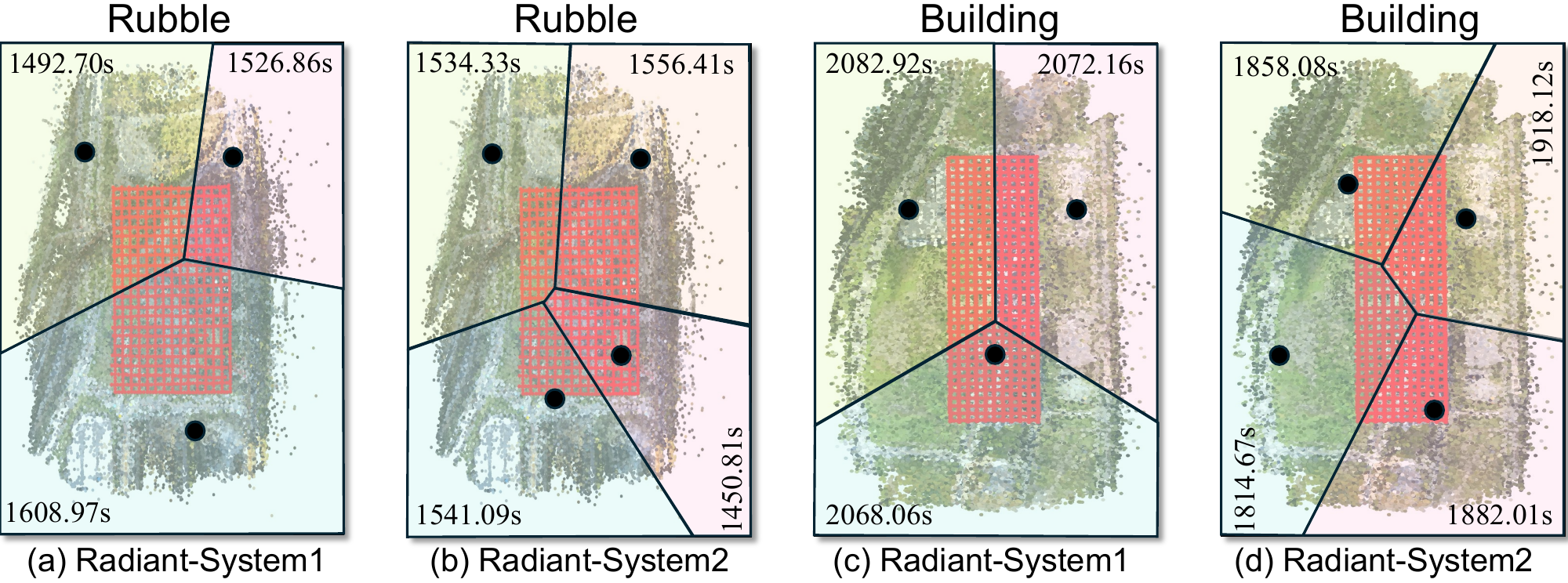}
\caption{The partitions of Radiant and the latency of each region.}
\label{fig:partition}
\end{center}
\vspace{-3mm}
\end{figure}

Furthermore, we demonstrated the partitioning results of the ARP algorithm under the settings of System 1 and System 2 in Fig.~\ref{fig:partition}, as well as the latency on each edge. 
This figure illustrates that the ARP algorithm can effectively balance the workload distribution based on the computational resources of each region.

\subsection{Deep Dive into Radiant}
\begin{figure}[!t]
\begin{center}
\includegraphics[width=\linewidth]{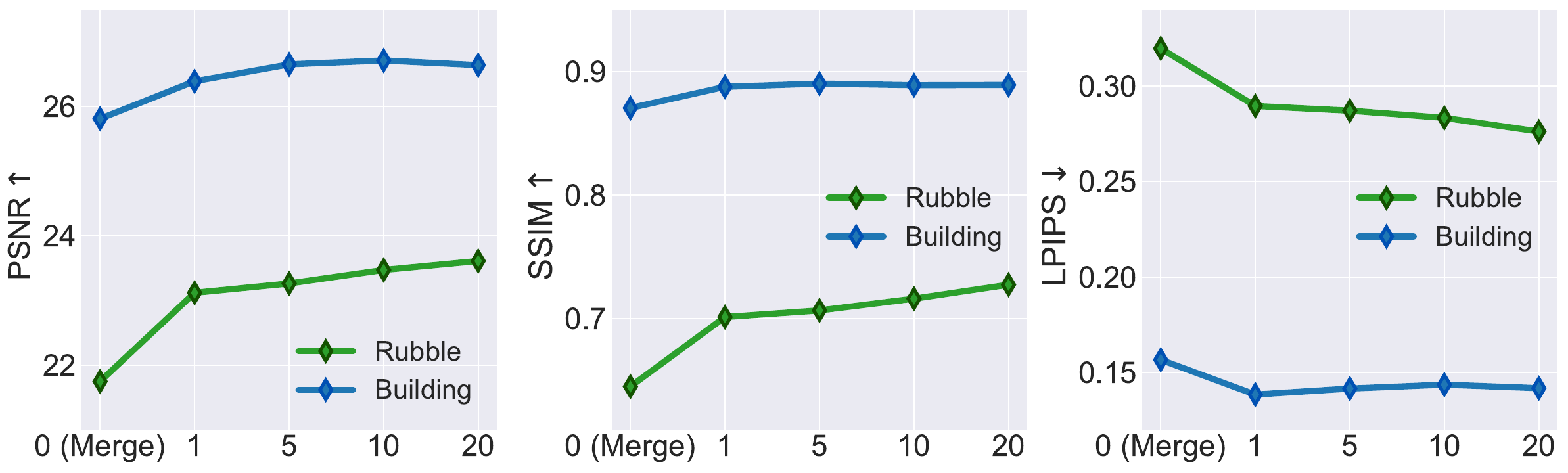}
\caption{The impact of the number of aggregate epochs on the quality of regional boundaries.}
\label{fig:fusion}
\end{center}
\end{figure}

We conducted ablation studies on the model's aggregation methods. 
Table~\ref{tab:fusion} shows a performance comparison between the model aggregation algorithm and directly merging according to the region's boundaries. 
It is evident that the aggregated model maintains better accuracy across three metrics. 
In terms of the selection of retaining epochs, Fig.~\ref{fig:fusion} demonstrates the impact of different training epochs on the quality of regional boundaries.
We find that using only a small number of epochs can already achieve much better results than the merging method.

\begin{figure}[!t]
\begin{center}
\includegraphics[width=\linewidth]{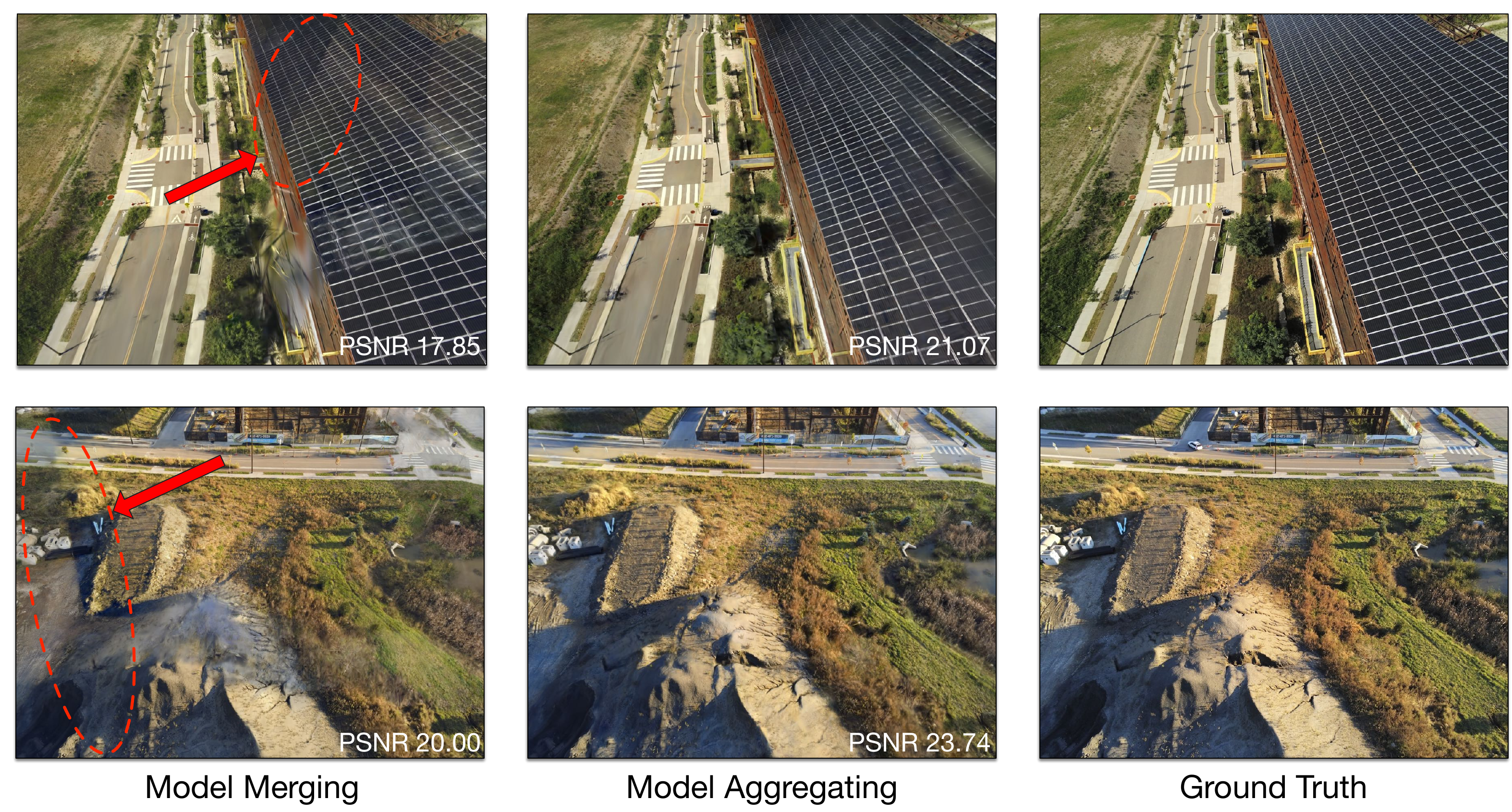}
\caption{The visualizations of two model fusion methods. The merging approach results in noticeable discontinuities at the boundaries.}
\label{fig:visual}
\end{center}
\end{figure}

The differences between these merging and aggregation methods are more visually apparent, as shown in Fig.~\ref{fig:visual}.
We displayed the effects of two fusion methods with images from two different datasets.
It was observed that using the merging method results in noticeable boundary lines, whereas the aggregation algorithm significantly alleviates this phenomenon.

\begin{figure}[!t]
\begin{center}
\includegraphics[width=0.8\linewidth]{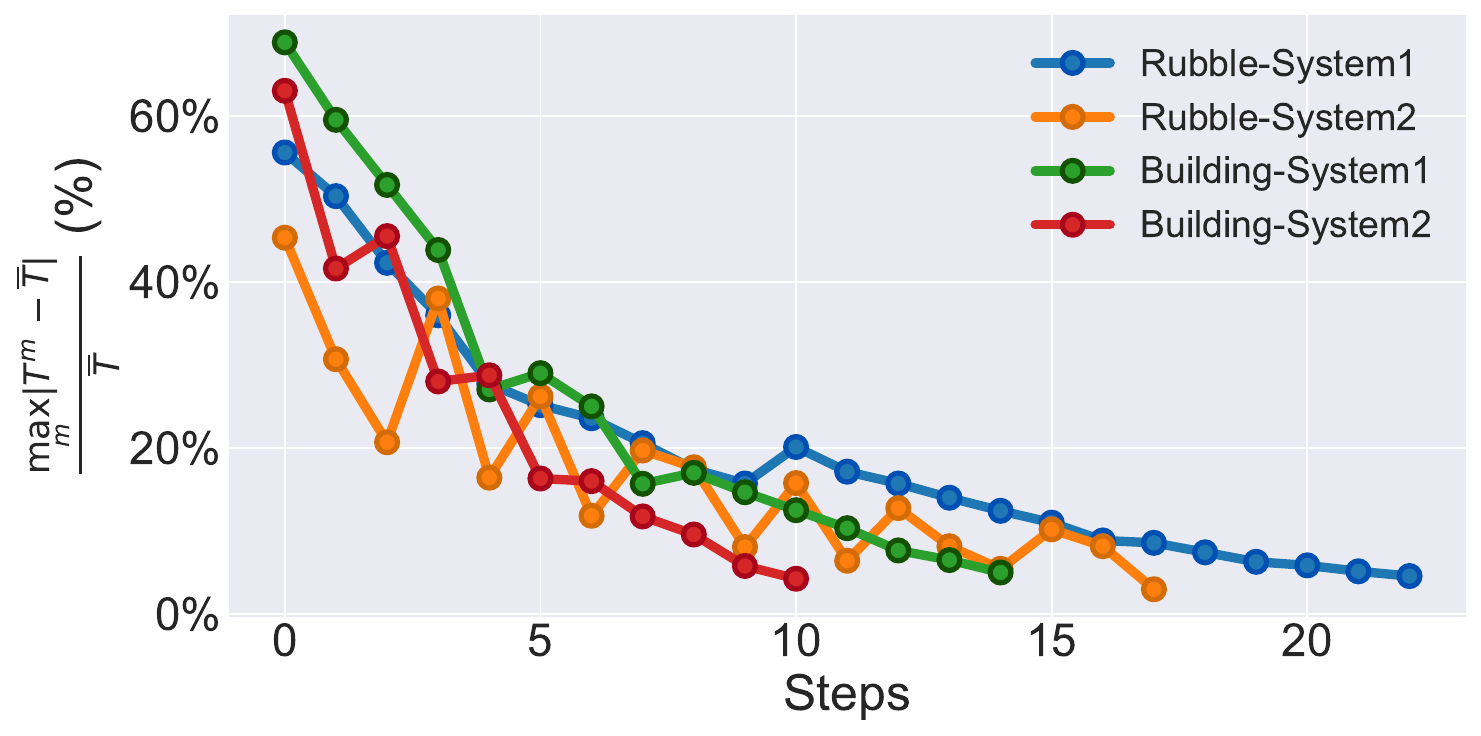}
\caption{The convergence process of the ARP algorithm.}
\label{fig:convergence}
\end{center}
\end{figure}

We also display the convergence process of the ARP algorithm in Fig.~\ref{fig:convergence}.
The y-axis represents the ratio of the maximum time difference in the current partitioned region to the average time, i.e., $(\max_m{|T^m-\overline{T}|})/{\overline{T}}$. 
A smaller value indicates a more balanced workload distribution.
Our ARP algorithm can rapidly converge after a small number of iterations.

\begin{table}[t]
\caption{Fusion methods ablation experiment.}
\label{tab:fusion}
\resizebox{0.49\textwidth}{!}{
\begin{tabular}{@{}c|c|ccc|ccc@{}}
\toprule
\multirow{2}{*}{\textbf{Settings}} &
  \multirow{2}{*}{\textbf{Method}} &
  \multicolumn{3}{c|}{\textbf{Rubble}} &
  \multicolumn{3}{c}{\textbf{Building}} \\ \cmidrule(l){3-8} 
 &
   &
  \multicolumn{1}{c|}{\textbf{PSNR}$\uparrow$} &
  \multicolumn{1}{c|}{\textbf{SSIM}$\uparrow$} &
  \textbf{LPIPS}$\downarrow$ &
  \multicolumn{1}{c|}{\textbf{PSNR}$\uparrow$} &
  \multicolumn{1}{c|}{\textbf{SSIM}$\uparrow$} &
  \multicolumn{1}{c|}{\textbf{LPIPS}$\downarrow$} \\ \midrule
\multirow{2}{*}{\textbf{\begin{tabular}[c]{@{}c@{}}System1\end{tabular}}} &
  \textbf{Merge} &
  \multicolumn{1}{c|}{22.51} &
  \multicolumn{1}{c|}{0.637} &
  0.371 &
  \multicolumn{1}{c|}{21.17} &
  \multicolumn{1}{c|}{0.709} &
  \multicolumn{1}{c|}{0.278} \\
 &
  \textbf{Aggregate} &
  \multicolumn{1}{c|}{24.44} &
  \multicolumn{1}{c|}{0.737} &
   0.305&
  \multicolumn{1}{c|}{21.57} &
  \multicolumn{1}{c|}{0.716} &
  \multicolumn{1}{c|}{0.292} \\ \midrule
\multirow{2}{*}{\textbf{\begin{tabular}[c]{@{}c@{}}System2\end{tabular}}} &
  \textbf{Merge} &
  \multicolumn{1}{c|}{23.00} &
  \multicolumn{1}{c|}{0.682} &
  \multicolumn{1}{c|}{0.328} &
  \multicolumn{1}{c|}{20.46} &
  \multicolumn{1}{c|}{0.658} &
  \multicolumn{1}{c|}{0.310} \\
 &
  \textbf{Aggregate} &
  \multicolumn{1}{c|}{24.53} &
  \multicolumn{1}{c|}{0.740} &
  \multicolumn{1}{c|}{0.284} &
  \multicolumn{1}{c|}{21.66} &
  \multicolumn{1}{c|}{0.723} &
  \multicolumn{1}{c|}{0.236} \\ \bottomrule
\end{tabular}}
\end{table}

\section{Related Work}\label{sec6}
In this section, we briefly review the latest work on 3DGS technology and large-scale hierarchical frameworks developed in recent years.

\subsection{3D Scene Reconstruction}
Traditional approaches~\cite{agarwal2011building, fruh2004automated, schonberger2016structure} follow a structure-from-motion pipeline that estimates camera poses and generates sparse point clouds. However, such methods often contain artifacts or holes in areas with limited texture or speculate reflections as they are challenging to triangulate across images. 
Recently, NeRF~\cite{mildenhall2021nerf} and 3DGS~\cite{kerbl20233d} variants have become a worldwide 3D representation system thanks to their photo-realistic characteristics and the ability of novel-view synthesis, which inspires many works~\cite{zhenxing2022switch, zhang2023efficient, turki2022mega, tancik2022block, lin2024vastgaussian, xu2023grid, suzuki2024fed3dgs} to extend it into large-scale scene reconstructions. 
The above methods can be categorized into centralized and distributed frameworks. 
Centralized methods~\cite{xu2023grid,zhang2023efficient} adopt the integration of NeRF-based and grid-based methods to model city-scale reconstruction. 
Distributed methods~\cite{lin2024vastgaussian,turki2022mega} apply scene decomposition for multiple NeRF / Gaussian models optimization.
However, with the growing scene size, both centralized and distributed variants limit their scalability due to the central server's limited data storage and unacceptable computation costs. 
Nearly proposed Fed3DGS~\cite{suzuki2024fed3dgs} introduces federated learning to leverage the computational resources of clients and emerge clients' 3DGS models into the central server by a distillation update scheme.
Nevertheless, this method ignores the scales and the bundle adjustment between different edge models, resulting in a performance drop. Meanwhile, it only focuses on over-fitting the photo-realistic rendering but ignores the geometry performance.

\subsection{Large-Scale Hierarchical Framework}

The hierarchical cloud-edge-device framework~\cite{liu2020client,cui2022optimizing,deng2021share} has been proposed to address large-scale distributed edge computing problems and has been widely applied in various distributed domains in recent years. 
Wang et al.~\cite{wang2021resource} identified the optimal clustering configuration and implemented hybrid federated learning using a blend of synchronous and asynchronous methods.
%
%
Li et al. introduced FedGS~\cite{li2022data} for the IIoT environment, which utilizes the gradient-based binary permutation algorithm (GBP-CS) to select subsets.
%
%
Mhaisen et al.~\cite{mhaisen2021optimal} developed an optimization for the pairing of devices and edges to reduce the distance in class distribution.
Zhong et al.~\cite{zhong2022flee} proposed the FLEE algorithm, which leverages edges and devices to achieve dynamic model updates.
Deng et al.~\cite{deng2023hierarchical} proposed FedHKT, a hierarchical framework that utilizes a hybrid knowledge transfer mechanism to enhance learning efficiency and performance.
Yang et al. proposed HierMo\cite{yang2023hierarchical}, which applies momentum to accelerate distributed training.
Chen et al.~\cite{chen2023taming} introduced SD-GT, which enhances federated learning on fog networks by optimizing device updates and improving model performance.

\section{Conclusion}\label{sec7}
We designed Radiant, a hierarchical framework for 3D Gaussian rendering in large scene reconstruction with resource-heterogeneous systems. 
It is the first distributed framework that simultaneously satisfies efficiency, scalability, and privacy.
We propose the ARP and RTP algorithms to distribute the reconstruction workload among edges and devices. 
Additionally, we develop a novel model aggregation method for the hierarchical framework to enhance the model quality.
Finally, we develop a testbed and experiments demonstrate that Radiant outperforms the state-of-the-art methods on model quality and end-to-end latency.

\bibliographystyle{ieeetr}
\bibliography{infocom}

\end{document}